\documentclass[pdflatex,sn-mathphys-num]{sn-jnl}

\usepackage{graphicx}%
\usepackage{multirow}%
\usepackage{amsmath,amssymb,amsfonts}%
\usepackage{amsthm}%
\usepackage{mathrsfs}%
\usepackage[title]{appendix}%
\usepackage{xcolor}%
\usepackage{textcomp}%
\usepackage{manyfoot}%
\usepackage{booktabs}%
\usepackage{algorithm}%
\usepackage{algorithmicx}%
\usepackage{algpseudocode}%
\usepackage{listings}%

\theoremstyle{thmstyleone}%
\theoremstyle{thmstyletwo}%

\theoremstyle{thmstylethree}%

\begin{document}
\title[Article Title]{Language Markers of Emotion Flexibility Predict Depression and Anxiety Treatment Outcomes}

\author[1]{\fnm{Benjamin} \sur{Brindle}}

\author[2]{\fnm{George A.} \sur{Bonanno}}

\author[3]{\fnm{Thomas Derrick} \sur{Hull}}

\author[4]{\fnm{Nicolas} \sur{Charon}}

\author*[5,6]{\fnm{Matteo} \sur{Malgaroli}}\email{matteo.malgaroli@nyulangone.org}

\affil[1]{\orgdiv{Department of Applied Mathematics and Statistics}, \orgname{Johns Hopkins University}}

\affil[2]{\orgdiv{Department of Clinical Psychology}, \orgname{Columbia University}}

\affil[3]{\orgname{Talkspace}, \orgaddress{ \city{New York}, \state{New York}, \country{United States of America}}}

\affil[4]{\orgdiv{Department of Mathematics}, \orgname{University of Houston}}

\affil[5]{\orgdiv{Center for Data Science}, \orgname{New York University}}

\affil[6]{\orgdiv{Department of Psychiatry}, \orgname{New York University School of Medicine}}








\abstract{Predicting treatment non-response for anxiety and depression is challenging, in part because of sparse symptom assessments in real-world care. We examined whether passively captured, fine-grained emotions serve as linguistic markers of treatment outcomes by analyzing 12 weeks of de-identified teletherapy transcripts from 12,043 U.S. patients with moderate-to-severe anxiety and depression symptoms. A transformer-based small language model extracted patients' emotions at the talk-turn level; a state-space model (VISTA-SSM) clustered subgroups based on emotion dynamics over time and produced temporal networks. Two groups emerged: an improving group (n=8,230) and a non-response group (n=3,813) showing increased odds of symptom deterioration, and lower likelihood of clinically significant improvement. Temporal networks indicated that sadness and fear exerted most influence on emotion dynamics in non-responders, whereas improving patients showed balanced joy, sadness, and neutral expressions. Findings suggest that linguistic markers of emotional inflexibility can serve as scalable, interpretable, and theoretically grounded indicators for treatment risk stratification.}

\keywords{Flexibility, Depression, Anxiety, Small Language Models, Clustering, Network Analysis}



\maketitle

\section{Introduction}\label{sec:introduction}



Depression and anxiety represent a growing public health crisis with substantial emotional, social, and economic costs \cite{counts2023psychological}. As the demand for care rises, digital health interventions offer an accessible alternative to traditional in-person treatment \cite{pullmann2025message,hull2020two}. Yet despite its promise, many participants fail to benefit, as with other mental health interventions. Predicting non-response is often hampered by a lack of consistent symptom assessments, and, in response to this challenge, there is increasing interest in predicting treatment outcomes using behavioral digital markers \cite{doherty2025passive}. Computational models could perform a critical role in identifying individuals at risk of non-response, and informing the allocation or reallocation of scarce therapeutic resources. However, for these methods to also be clinically useful, they must be grounded in a theoretical framework that highlights interpretable treatment targets.

One promising avenue is the analysis of automatically-collected linguistic markers of emotional expression \cite{malgaroli2023natural}. Prior studies have demonstrated that early use of positive sentiment, change talk, linguistic distancing, and markers of behavioral activation are associated with greater symptom improvements \cite{Burkhardt2021,Ewbank2021,Shapira2022,Malgaroli2024,Nook2022}. Other work has explored changes in emotional expression over the course of treatment, finding that an increase in positive sentiment and a decrease in first-person singular words throughout therapy were associated with symptom reductions \cite{Shapira2022,Eberhardt2025}. However, much of this work is limited by relying on lexicon- or rule-based approaches such as LIWC and VADER \cite{Burkhardt2021,Nook2022,Shapira2022,Malgaroli2024}, which cannot capture  context-dependent meaning or linguistic nuance. Subsequent studies have applied deep learning models that can detect subtle emotional cues including recurrent neural networks that capture sequential dependencies (e.g., BiLSTM \cite{Ewbank2021}), and transformer-based approaches that capture long-range context (e.g. \cite{Eberhardt2025}). However, another key limitation across these studies is that they aggregate sentiment at the week level, a coarse temporal granularity that misses clinically meaningful emotional fluctuations occurring within treatment \cite{malgaroli2023natural,Luo2025,Paz2024}.

Parallel evidence from experience sampling and ecological momentary assessment (EMA) underscores the importance of short-timescale affect dynamics for treatment response. While some work suggests that baseline levels of positive and negative affect, or the ratio between them, are sufficient predictors of treatment response \cite{Forbes2012,Husen2016}, many studies indicate that emotional fluctuations carry additional predictive value. For example, Husen and colleagues \cite{Husen2016} found that fluctuations in negative affect pre-treatment predicted early treatment response, whereas mean affect levels did not. Other work has explored how positive affect interacts with daily regulatory processes, either by enhancing emotion regulation \cite{Hehlmann2024} or by suppressing subsequent negative affect \cite{Wichers2012}. In a complementary finding, Peeters and colleagues  \cite{Peeters2010} reported that lower emotional reactivity to both positive and negative daily events at baseline predicted greater depression symptom severity one month into treatment, suggesting that both the presence and functional significance of affective dynamics are important for treatment response. 

Part of the challenge lies in grounding investigations of emotion dynamics in theories that account for their complexity. Conventional perspectives on the role of emotion in treatment often fall prey to the fallacy of uniform efficacy \cite{bonanno2013regulatory}; that is, they typically frame painful emotions, like fear or anger, as indicators of dysfunction, while positive states, such as joy and happiness, are viewed as uniform markers of mental health. Although the simplicity of this supposition is appealing, in their immediate moment-to-moment manifestations, emotions are essentially functional such that variations in both positive and negative emotions can be adaptive or maladaptive depending on the context in which they occur \cite{coifman2016context,coifman2010distress,gruber2011dark}. A growing body of research has shown this to be the case as well for the strategic regulation of emotion \cite{bonanno2004importance,troy2013person} (for review see \cite{bonanno2023resilience}). Integrative theoretical models developed in response to this research, such as the regulatory flexibility framework \cite{bonanno2013regulatory,bonanno2021resilience}, provide a dynamic vehicle for understanding how individuals adaptively manage emotional experiences in the face of stress. This approach shifts attention from the simple presence of putatively healthy or unhealthy emotions to their skillful modulation in response to momentary shifts in situational demands. Regulatory flexibility thus highlights how balanced and varied emotional experience and expression may support adjustment, resilience, and therapeutic responsiveness. Critically, it motivates analytic methods that can capture dynamic relationships between emotions and adjustment. 

Taken together, transcript-based sentiment analysis and EMA studies both underscore the relevance of emotional dynamics for treatment outcomes, yet each approach has limitations: linguistic analyses typically aggregate across sessions, while EMA imposes heavy participant burden. This gap points to the need for methods that capture fine-grained emotional fluctuations directly. A further methodological challenge is that most longitudinal clustering algorithms are unable to handle data that is both unequally spaced over time and imbalanced in number of observations \cite{lu2024clustering}, necessitating specialized approaches to identify patient subgroups exhibiting distinct dynamic patterns and linking them to treatment outcomes.

\begin{figure}[H]
    \centering
    \includegraphics[width=1.0\linewidth]{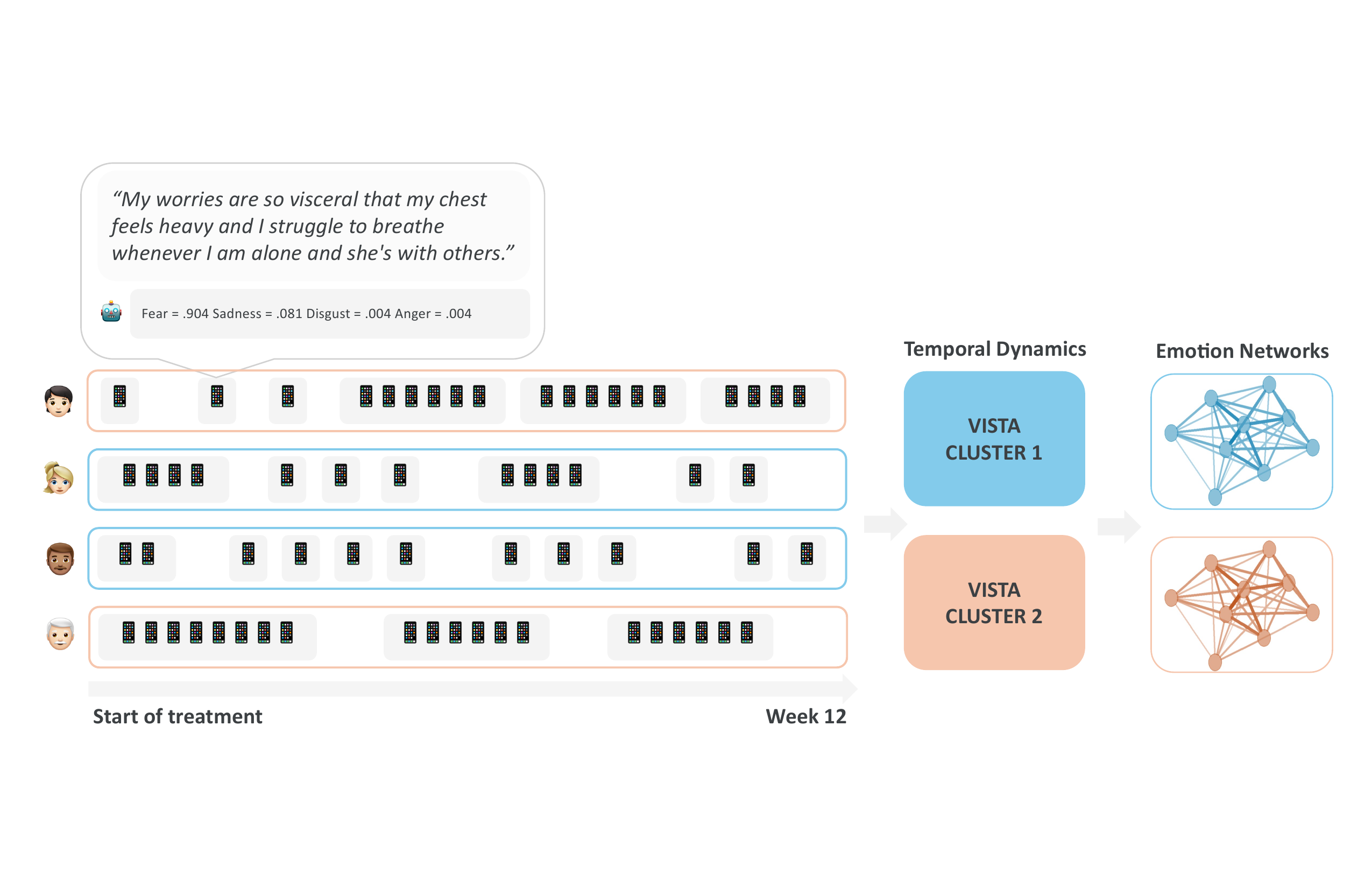}
    \caption{High level depiction of our methodology. Over the course of 12 weeks, patients participate in message-based psychotherapy. We infer patient emotions in each talk turn using a transformer-based language model. VISTA clusters the patients based on temporal dynamics of emotions. For each cluster, we create emotion networks to identify central emotions.}
    \label{fig:overall}
\end{figure}

In the current study, we leverage a large psychotherapy cohort of over twelve thousand patients to identify emotion dynamics over treatment associated with outcomes. We first use a Small Language Model (SLM) \cite{belcak2025small} to extract utterance-level patients' emotions from therapy transcripts. We then capture temporal flow across multiple emotions over treatment using VISTA-SSM (Varying and Irregular Sampling Time-series Analysis via State Space Models, see \cite{brindle2024vista}), and test whether resulting clusters of emotional dynamics are prospectively associated with depression and anxiety outcomes. Finally, we construct cluster-specific temporal networks to interpret the underlying emotional dynamics and identify emotions with greater causal influence within each network, as shown in figure \ref{fig:overall}. By linking early patterns of emotional expressions to subsequent treatment response, this approach could support risk stratification and adaptive allocation of treatment resources.

\section{Methods}\label{sec:methods}

\subsection{Participants and Setting}\label{ssec:participants}

The study sample comprised individuals in the United States seeking mental health treatment who enrolled in a digital mental teletherapy platform (www.talkspace.com) utilized by independent licensed clinicians. Access to this platform is available through employee assistance programs, direct self-enrollment, and coverage under individual insurance behavioral health benefits. The study consisted of individuals engaged in message-based psychotherapy \cite{pullmann2025message} through the platform between 2016 and 2021. Prior to platform use, all patients and clinicians provided written consent to the use of their de-identified data for research purposes. Study procedures were approved by the NYU School of Medicine IRB (i23-00345). 

\subsubsection{Sample}\label{sssec:sample}

Study eligibility required patients to have been diagnosed with depression or anxiety by their assigned licensed mental health provider and to have clinically significant symptoms at baseline, defined as scoring 10 or higher on the Patient Health Questionnaire-9 (PHQ-9) for depression and/or the Generalized Anxiety Disorder-7 (GAD-7) for anxiety \cite{kroenke2001phq,kroenke2007anxiety}. Exclusion criteria consisted of comorbid psychosis, bipolar, and autism, as these conditions present unique clinical trajectories, or therapy transcripts deviating from symptoms assessment and typical engagement patterns.

The initial sample consisted of 131,707 patients who had cumulatively completed 278,254 assessments. First, we removed patients with comorbid diagnoses of psychosis, bipolar disorder, schizophrenia, autism, Asperger’s, dementia, and factitious disorders, yielding 122,660 patients. Next, we removed patients who had no associated psychotherapy transcript files. To center the transcripts around completion of the assessments, we also removed patients who had gaps of more than four weeks or less than two weeks between any two of their first five assessments, disregarding this if they had fewer than five assessments total. These exclusions gave a sample of 32,315 patients. We then removed patients who had transcripts with non-English language messages and who have not interacted with their therapists between any two assessments to reduce the sample to 32,176 patients. As our analysis aimed to examine the evolution of clinically significant anxiety and depression symptoms, we restricted our attention to the 22,954 patients who had baseline GAD-7 or PHQ-9 score of at least 10. To ensure consistent platform engagement, we included only patients who completed at least 20 talk turns, yielding our final sample of 12,043 patients.

\subsection{Transcripts and Assessments}\label{ssec:assessments}

Patients were assigned to online ``rooms'' where they communicated with a licensed clinician. Patients could message their provider during office hours, typically 9 AM to 6 PM in the provider's time zone, five days per week. For further details on message-based therapy, see Pullmann et al. \cite{pullmann2025message}). We focused on the initial twelve weeks of treatment, consistent with psychotherapy standards for early treatment evaluation and response monitoring. Psychotherapy transcripts were generated and de-identified from the room logs by removing any personal identifiers, proper nouns, locations and dates.

Patients completed symptoms assessments approximately every three weeks, filling out both the PHQ-9 and GAD-7 to provide a measure of their anxiety and depression symptoms over the prior two weeks. Patients first completed a baseline assessment, then again at approximately weeks 3, 6, 9, and 12. Some patients dropped out following weeks 3, 6, and 9. Responses on all items on the PHQ-9 and GAD-7 are given on a 4-point Likert scale (0 = Not at all to 3 = Nearly every day) and ask how often the patient has experienced specific problems within the previous 2 weeks. The transcripts were analyzed within this framework by first dividing them into talk turns between patient and provider. For the patient, this means binning all texts they sent the provider in between any two messages sent by the provider. We anchored the talk turns to the symptoms assessments by considering only those talk turns which occurred after the previous assessment but not before the 3 weeks prior to the current assessment in an attempt to capture the format of the questions inquiring about symptoms over the previous 2 weeks. 

\subsection{Data analysis}\label{ssec:analyses}

All analysis were implemented in Python. Code for performing the clustering and constructing the temporal networks is available online (\url{https://github.com/benjaminbrindle/vista_ssm}) under an open-source license.

\subsubsection{Emotions Analysis}\label{sssec:transcripts}

Psychotherapy transcripts were segmented into discrete patient talk turns, defined as all messages sent from a patient before a response from their provider. Each talk turn could be of any length and was analyzed separately. In order to detect emotions at the talk turn level, we leveraged the SLM DistilRoBERTa-base \cite{hartmann2022emotionenglish}, which fine tunes the encoder-based transformer model RoBERTa \cite{liu2019roberta} to identify emotions in English text data. This model has been applied in psycholinguistic analyses \cite{butt2022goes} and research examining the relationship between sentiment and mental health \cite{pathak2025sentiment}. We used DistilRoBERTa-base to transform each talk turn into a vector of length seven, where each element ranges $[0,1]$ representing the strength in the text of a specific emotion (anger, disgust, fear, joy, neutral, sadness, and surprise).

\subsubsection{Longitudinal Clusters of Emotions}\label{sssec:vista}

We performed model-based longitudinal clustering of talk-turn emotion vectors over the 12-week treatment period using VISTA \cite{brindle2024vista}. VISTA enables simultaneous parametric modeling and clustering of multivariate time series data with irregular and varying sampling rates, making it suitable for our scenario, as the number of talk-turns and their temporal frequency varied across patients. VISTA demonstrated superior performance compared to classical clustering methods and, when applied to psychological data, produced unsupervised clusters that corresponded with mental health information withheld from the model, supporting its use in situations without known ``ground truth'' labels for patient time series \cite{brindle2024vista}.

We provide a brief overview of VISTA. Given a set of $N$ unlabelled time series $Y^i = (y^i_k)_{k=1,...,T_i}$ irregularly sampled at times $t_1,\dots,t_{T_i}$ where $T_i$ denotes the number of samples for time series $i$, VISTA automatically assigns each of these to a certain group (or cluster), with the number $M$ of those groups being predefined. The determination of the clusters and the cluster assignment relies on a generative model whereby each time series in a specific cluster is assumed to arise as a realization of a linear stochastic dynamical system known as a linear Gaussian state space model \cite{ho1964bayesian} of the form:
\begin{align}
    x_k &= A_{l,k} x_{k-1} + w_{l,k} \nonumber \\
    y_k &= C_lx_k  + v_{l,k} \label{eq:lgssm}\\
    x_1 &= \mu_l + u_l. \nonumber
\end{align}
in which $l$ denotes the cluster to which the time series corresponds, $x_k$ is an underlying latent process generating the observations, and $A_{l,k} = \text{Id} + \Delta_k A_l$. Moreover, $\{w_{l,k}\}$, $\{v_{l,k}\}$, and $u_l$ are all Gaussian independent and identically distributed random variables with zero mean and respective covariance matrices $\Gamma_{l,k} = \Delta_k \Gamma_l$, $\Sigma_{l,k} = \frac{1}{\Delta_k}\Sigma_l$ and $P_l$, respectively, where $\Delta_k = t_{k} - t_{k-1}$. Therefore each cluster $l=1,\ldots,M$ is represented by the parameter set $\theta_l = \{\mu_l,A_l,C_l,P_l,\Sigma_l,\Gamma_l\}$. The whole model behind VISTA can be then interpreted as an extension of the mixture of LGSSMs approach of \cite{umatani2023time}. More specifically, VISTA jointly and iteratively estimates both cluster assignments and cluster parameters based on the expectation-maximization (EM) algorithm. Upon convergence of the EM scheme, VISTA returns the cluster membership for each time series and the parameter set $\theta_l = \{\mu_l,A_l,C_l,P_l,\Sigma_l,\Gamma_l\}$ for each cluster $l=1,\dots,M$. These parameters provide interpretability to the estimated clusters as they can be leveraged to simulate a ``typical'' trajectory for each group by taking \eqref{eq:lgssm} without noise terms removed. For a more in depth description of the approach, see Brindle et al. \cite{brindle2024vista}.   

We employed VISTA to identify heterogeneous clusters based on the trajectories of seven emotions obtained from patient talk-turn over twelve weeks of treatment. VISTA analyzed emotion vectors and their associated timestamps to produce unsupervised cluster assignments, along with a parameterized model for each cluster that describes the temporal evolution of sentiment. We examined baseline symptom-level differences in anxiety and depression symptoms between clusters using the Mann-Whitney U test \cite{mann1947test} and adjusted p-values for multiple testing using Bonferroni correction \cite{bonferroni1936teoria}. Employing logistic regression, we then associated patient-level cluster membership with clinical outcomes over 12 weeks of treatment. We defined patient outcomes of clinically significant change \cite{jacobson1992clinical} in either PHQ-9 or GAD-7 as moving below the clinical threshold (score of $<10$) and improving at least 5 points and deterioration as the worsening of symptoms by 5 or more points.

\subsubsection{Temporal Networks of Emotions}\label{sssec:network}
After defining clusters of temporal dynamics among the seven emotions and their association with clinical improvement and non-response, we identified the most central emotions driving these dynamics based on psychological network theory \cite{borsboom2021network}. We estimated temporal networks from patients' talk-turn emotions dynamics as a second step of the VISTA clustering procedure. Notably, characterizing how observed variables influence one another differently across clusters and throughout time evolution is a more challenging task than simply defining their longitudinal course. From \eqref{eq:lgssm}, we observe that the system does not directly relate $y_k$ and $y_{k-1}$, but rather models interactions at the level of the latent variables via the matrices $A_{l,k}$. In order to approximate the influence of $y_{k-1}$ on $y_k$, we observe that $x_k \approx A_{l,k} x_{k-1}$ and $y_k \approx C_l x_k$, yielding $y_k \approx C_l A_{l,k} C^{+}_l y_{k-1}$. $C^{+}_l$ is the Moore-Penrose pseudoinverse \cite{ben2003generalized}, used because $C_l$ is only a square matrix when $x$ and $y$ are of equal dimension, while in practice, a latent variable with higher dimension than the observed typically gives more meaningful results. This approximation allows us to consider the evolution of the observed variable with a simple linear equation for each cluster, shown in \eqref{eq:CAC}:
\begin{equation}
    y_k \approx C_l A_{l,k} C^{+}_l y_{k-1} = C_l(\text{Id} + \Delta_k A_l) C_l^{+} y_{k-1} \label{eq:CAC}
\end{equation}
For analysis at the cluster level, rather than at the individual level, we fix $\Delta_k$, which leads to a linear update equation relating $y_{k-1}$ to $y_k$ independent of $k$. In our setting, we set $\Delta_k$ to represent one week due to the assessment period being broken into three week intervals and the assessments themselves phrasing questions in terms of weeks. As we are interested in learning how each individual emotion in the talk turn influences the time evolution of the others throughout the clusters of patients returned by VISTA, we construct a temporal network from the matrix $C_l(\text{Id} + \Delta_k A_l) C_l^{+}$. From the temporal network, we gain insight into which specific emotions dynamic expressed in patient transcripts are associated with clinical outcomes by examining the centrality of each emotion \cite{borsboom2021network}. Specifically, we examine the Expected Influence \cite{robinaugh2016identifying} in the longitudinal network and compare centrality rankings \cite{malgaroli2021networks} of the most central emotions in each temporal network.

\section{Results}\label{sec:results}

\subsection{Sample Characteristics}\label{ssec:characteristics}

Demographic characteristics are provided in Table \ref{tab:demographics}. Patients in our sample had a mean treatment length of 126.3 (SD = 137.8) days with median 90 days. The transcripts in our sample had a median number of 34 talk turns, with a mean of 36.63 (SD = 13.22). Mean PHQ scores were 13.6 (SD = 5.2) at baseline, 10.1 (SD = 5.8) at week 3, 9.3 (SD = 6.0) at week 6, 8.9 (SD = 5.9) at week 9, and 8.5 (SD = 5.9) at week 12. GAD scores were 14.2 (SD = 4.5) at baseline, 10.4 (SD = 5.2) at week 3, 9.4 (SD = 5.3) at week 6, 9.0 (SD = 5.3) at week 9, and 8.6 (SD = 5.3) at week 12. A clinically significant improvement in symptom severity was observed in 33.0\% of participants for depressive symptoms (n = 3,972) and in 37.7\% for anxiety symptoms ( n = 4,543). Worsening symptoms were observed in 20.3\% of participants (n = 2441) on the PHQ-9 and in 18.2\% (n = 2186) on the GAD-7.

\begin{table}[h]
\centering
\caption{Demographics of patients (n=12,043) in our final sample.}
\label{tab:demographics}
\begin{tabular}{ll}
\hline \\
\begin{minipage}[t]{0.48\textwidth}

\begin{tabular}{p{0.75\textwidth}p{0.25\textwidth}}
Variable & \textbf{n (\%)}
\end{tabular}
\end{minipage}
&
\begin{minipage}[t]{0.48\textwidth}

\begin{tabular}{p{0.7\textwidth}p{0.3\textwidth}}
Variable & \textbf{n (\%)}
\end{tabular}
\end{minipage}
\\
\hline \\
\begin{minipage}[t]{0.48\textwidth}
\centering
\begin{tabular}{lr}

\textbf{Diagnoses} & \textbf{n = 12043} \\
Unspecified Anxiety Disorder & 3356 (27.8) \\
Major Depressive Disorder & 3183 (26.4) \\
Adjustment Disorder & 2294 (19.1) \\
Generalized Anxiety Disorder & 2214 (18.4) \\
Unspecified Mood Disorder & 409 (3.4) \\
Panic Disorder & 341 (2.8) \\
Dysthymia & 168 (1.4) \\
Social Anxiety Disorder & 78 (0.6) \\[1ex]

\textbf{Age Group}       & \textbf{n = 11305} \\
26--35 &  5757 (50.9) \\
18--25 &  3278 (29.0) \\
36--49 &  1906 (16.9) \\
50+    &   364 (3.2) \\[1ex]

\textbf{State}           & \textbf{n = 10979} \\
California       &  1422 (13.0) \\
New York         &  1129 (10.3) \\
Texas            &   785 (7.2) \\
Florida          &   696 (6.3) \\
Pennsylvania     &   428 (3.9) \\
Illinois         &   380 (3.5) \\
Washington       &   368 (3.4) \\
Ohio             &   350 (3.2) \\
Virginia         &   346 (3.2) \\
New Jersey       &   330 (3.0) \\
Other US states  &  4745 (43.2) \\

\end{tabular}
\end{minipage}
&
\begin{minipage}[t]{0.48\textwidth}
\centering
\begin{tabular}{lr}

\textbf{Ethnicity}       & \textbf{n = 1743} \\
Caucasian         & 1046 (60.0) \\
African American  &  229 (13.1) \\
Other             &  197 (11.3) \\
Asian             &  142 (8.1) \\
Hispanic          &  129 (7.4) \\[1ex]

\textbf{First Time in Therapy} & \textbf{n = 8452} \\
Yes        & 4925 (58.3) \\
No         & 3527 (41.7) \\[1ex]

\textbf{Gender}                & \textbf{n = 9848} \\
Female            & 7695 (78.1) \\
Male              & 1986 (20.2) \\
Non-binary/other  &  167 (1.7) \\[1ex]

\textbf{Education Level}       & \textbf{n = 5898} \\
Bachelor Degree or Higher  & 3583 (60.7) \\
High School                &  900 (15.3) \\
Some College               &  495 (8.4) \\
Masters Degree             &  479 (8.1) \\
Associates Degree          &  194 (3.3) \\
Doctoral Degree            &  105 (1.8) \\
Professional Degree        &   77 (1.3) \\
Less than high school      &   65 (1.1) \\[1ex]

\textbf{Marital Status}        & \textbf{n = 9075} \\
Single                      & 3779 (41.6) \\
Married                     & 2702 (29.8) \\
In Relationship/Cohabiting  & 2114 (23.3) \\
Previously Married          &  480 (5.3) \\

\end{tabular}
\end{minipage}
\\
\hline
\end{tabular}
\end{table}

\subsection{Longitudinal Clusters of Emotions }\label{ssec:vistatranscripts}

\begin{figure}[H]
    \centering
    \includegraphics[scale=0.35]{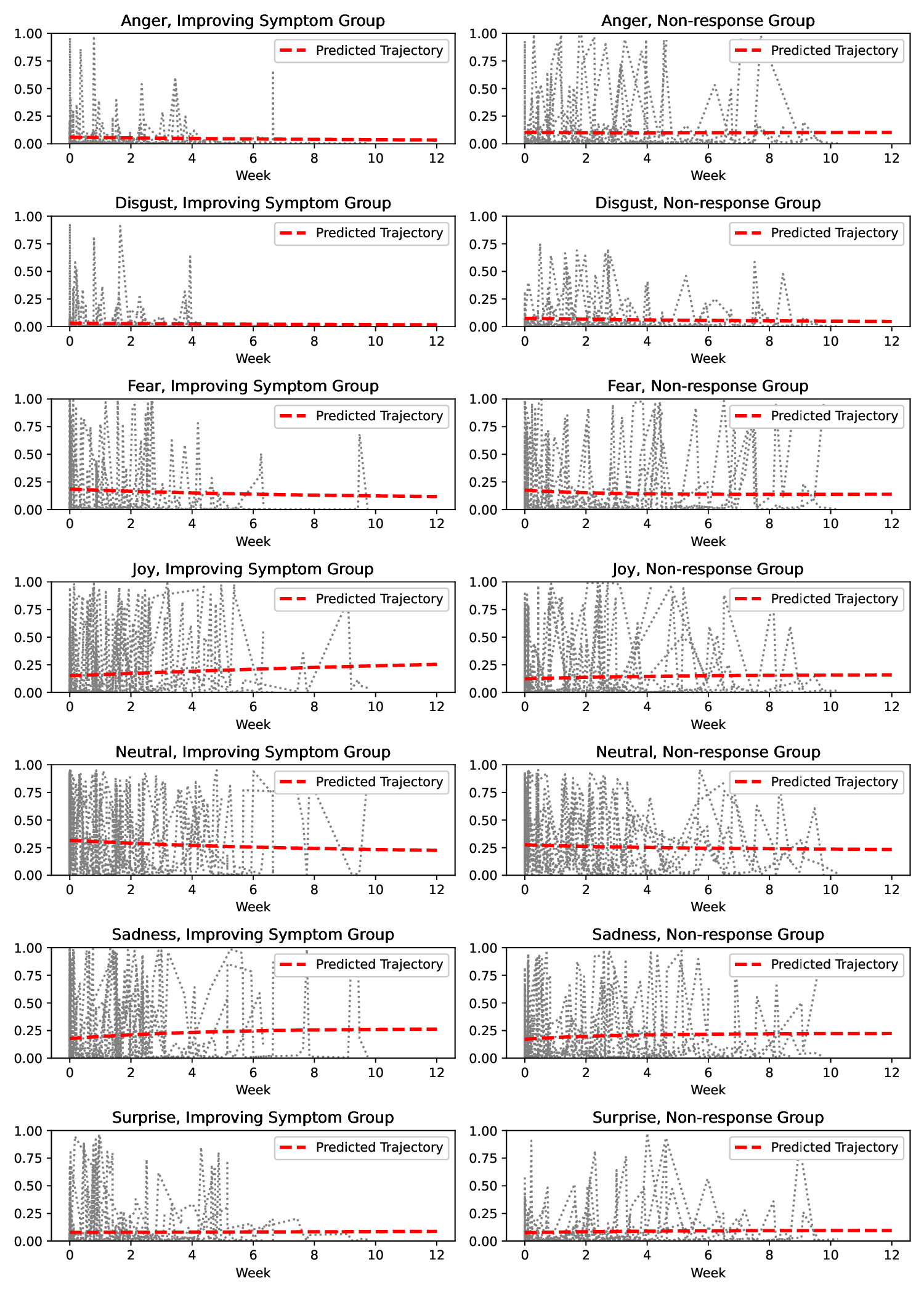}
    \caption{The result of VISTA with the returned clusters and noiseless trajectory (in red) determined by the optimal parameters along with randomly selected time series from each cluster. The x-axis is in week, while the y-axis is the emotion score in $[0,1]$ returned by DistilRoBERTa-base.}
    \label{fig:vistaplot}
\end{figure}

We applied the VISTA temporal clustering approach described in the methods to the entire sample, using $M=2$ clusters for parsimony. Parameters were initialized with VISTA's identity initialization scheme described in \cite{brindle2024vista} with a latent dimension of seven, matching the seven observed features. In figure \ref{fig:vistaplot}, we plot the noiseless trajectories of the different emotion features predicted from the estimated VISTA parameters associated to each cluster. Visually examining the noiseless trajectories, we see that the negative emotions of anger and disgust are typically higher within the second cluster (n = 3813, 31.66\%), while joy is higher in the first (n = 8230, 68.34\%). 

\begin{figure}[H]
    \centering
    \includegraphics[width=\linewidth]{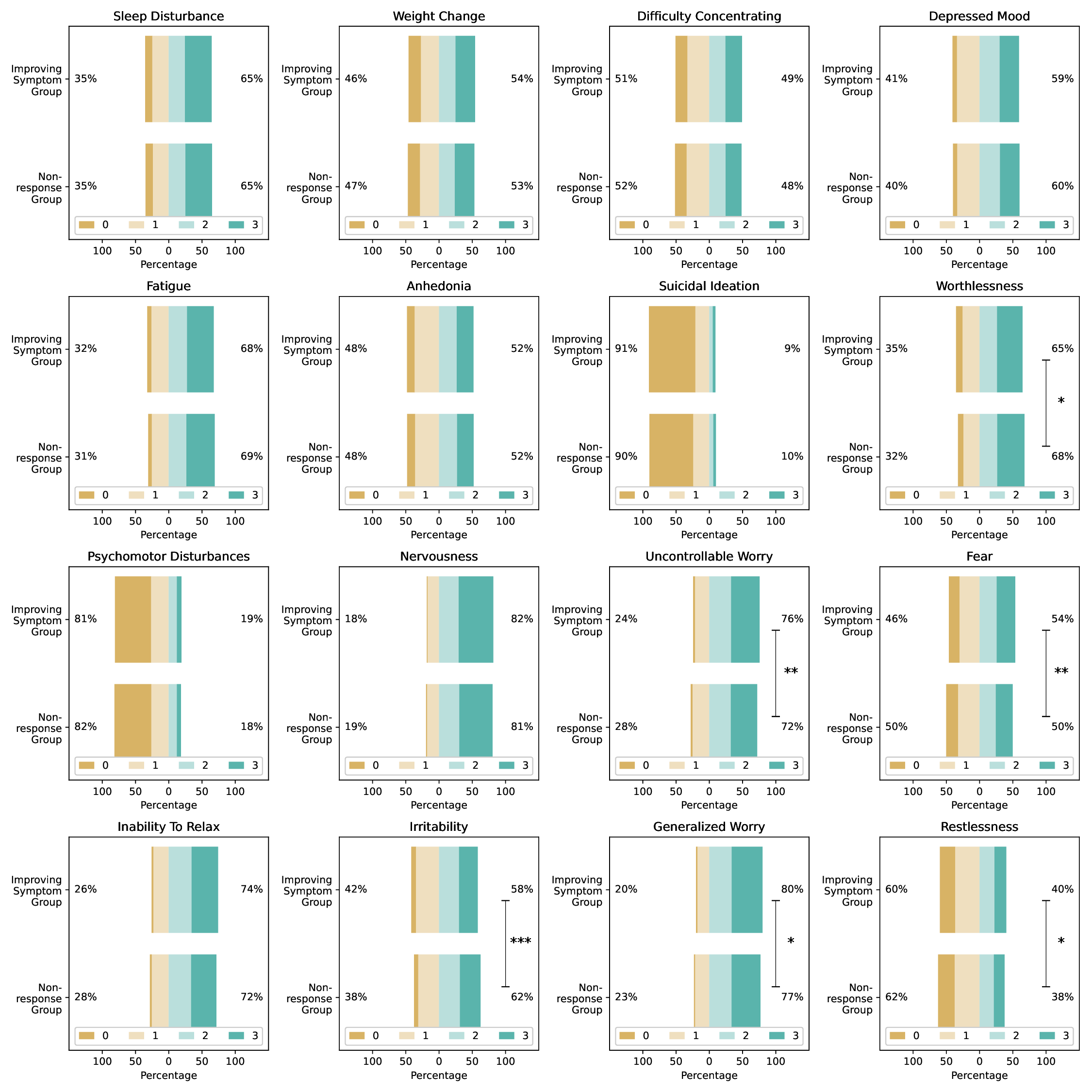}
    \caption{Likert plots of initial symptoms differences between emotion clusters in Improving (Cluster 1) and Non-response (Cluster 2) groups. Significant difference  between the two clusters, as measured by the Mann-Whitney U test, are denoted with asterisks (*), with *** indicating $p < 0.0001$, ** $p < 0.001$, and * $p < 0.05$, where $p$ is the Bonferroni-corrected p-value. Percentages indicate patients in the cluster below (left) or above (right) the clinical threshold (2+) for that symptom.}
    \label{fig:likert}
\end{figure}

Figure \ref{fig:likert} shows baseline symptoms from PHQ-9 and GAD-7 items for each patient by cluster membership. Mann-Whitney U tests indicated that the first cluster reported higher clinical levels of worry (generalized: 80\%, n = 6589; uncontrollable: 76\%, n = 6226), fear (54\%, n = 4423), and restlessness (40\%, n = 3312) compared to the second cluster. In turn, the second cluster endorsed higher levels of baseline worthlessness (68\%, n = 2575) and irritability (62\%, n = 2382). Full results and Bonferroni-corrected p-values are reported in the Supplementary Materials.

To determine clinical outcomes differences by twelve weeks of treatment, we ran logistic regression associating VISTA cluster membership with clinical significant change and deterioration for both anxiety and depressive symptoms, while controlling for age, gender, and education. Results indicated that the second cluster was more likely to exhibit deteriorating symptoms, as measured by both PHQ-9 scores (odds ratio = 1.127, 95\% CI  $[1.026, 1.239]$, p = .013) and GAD-7 scores (odds ratio = 1.145, 95\% CI  $[1.036, 1.264]$, p = .008), and less likely to exhibit a clinically significant change in anxiety symptoms (odds ratio = 0.92, 95\% CI  $[0.843, 0.994]$, p = .03) and depression symptoms (odds ratio = 0.872, 95\% CI  $[0.803, 0.947]$, p = .001) compared to the first VISTA cluster. We will therefore designate the first cluster as the \textit{improving symptom group}, while the second as the \textit{non-response group}.

\subsection{Temporal Networks of Emotions}\label{ssec:networkvista}

\begin{figure}[H]
    \centering
    \includegraphics[width=\linewidth]{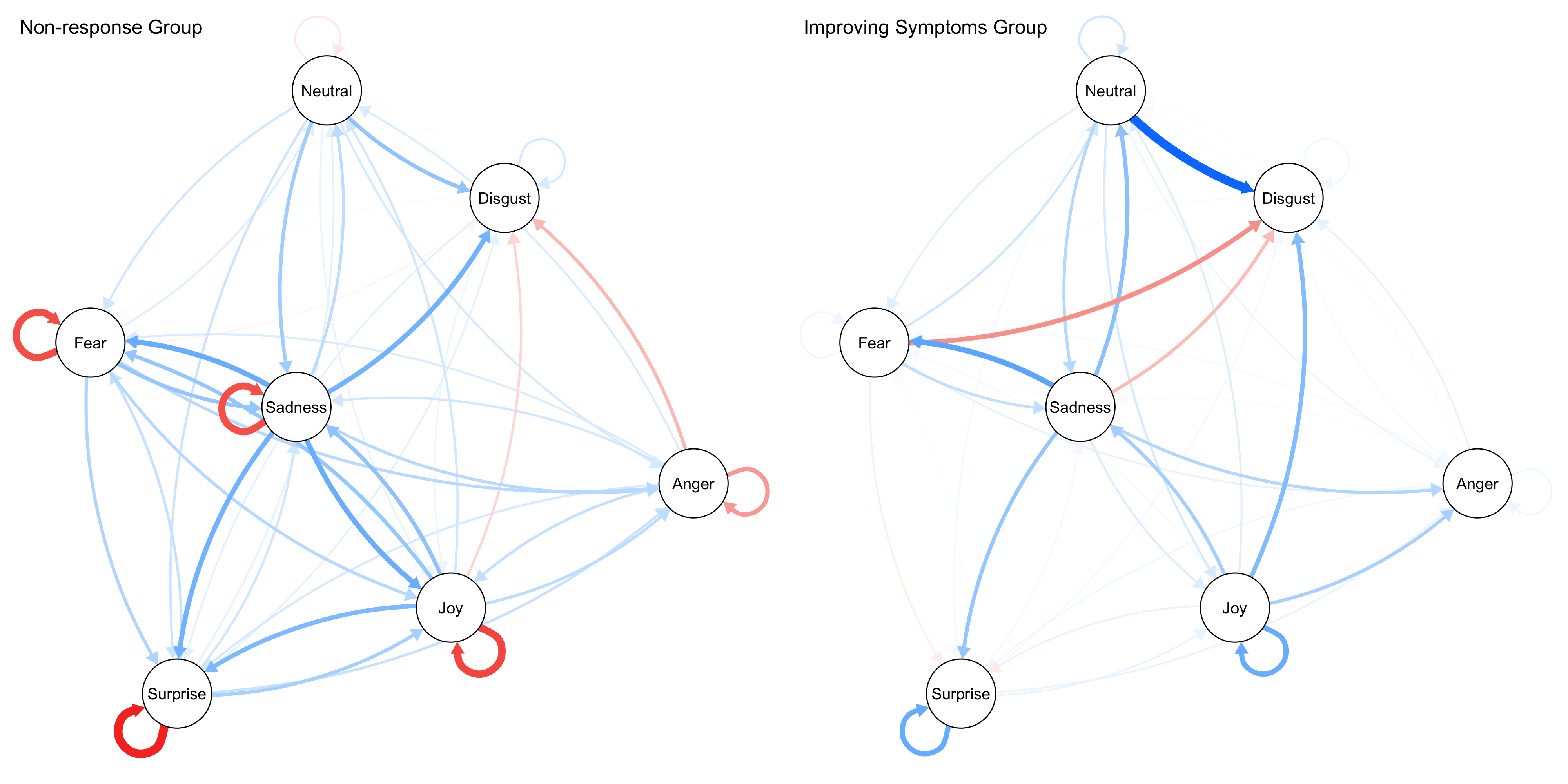}
    \caption{Temporal networks for both of the clusters returned by VISTA. Nodes represent emotions extracted from patient transcripts, while edges represent predictive temporal connections among symptoms. Blue edges represent positive association, with red representing negative association. Thicker edges represent stronger associations.}
    \label{fig:networks}
\end{figure}

To further examine cluster dynamics at the emotion level and relate these findings to interpret clinical outcome differences, we employed network analysis. Following the approach outlined in Section \ref{sssec:network}, the network is constructed from the matrix $C_l(\text{Id} + \Delta_k A_l) C_l^{+}$. We first examined the influence of emotions on one another as they evolve through time. In Figure \ref{fig:networks}, we observe that in the improving symptom group, joy and surprise increase autoregressively, while in the non-response group, nearly all emotions show autoregressive decreases, with sadness amplifying other emotions. 

\begin{figure}[H]
    \centering
    \includegraphics[scale=0.45]{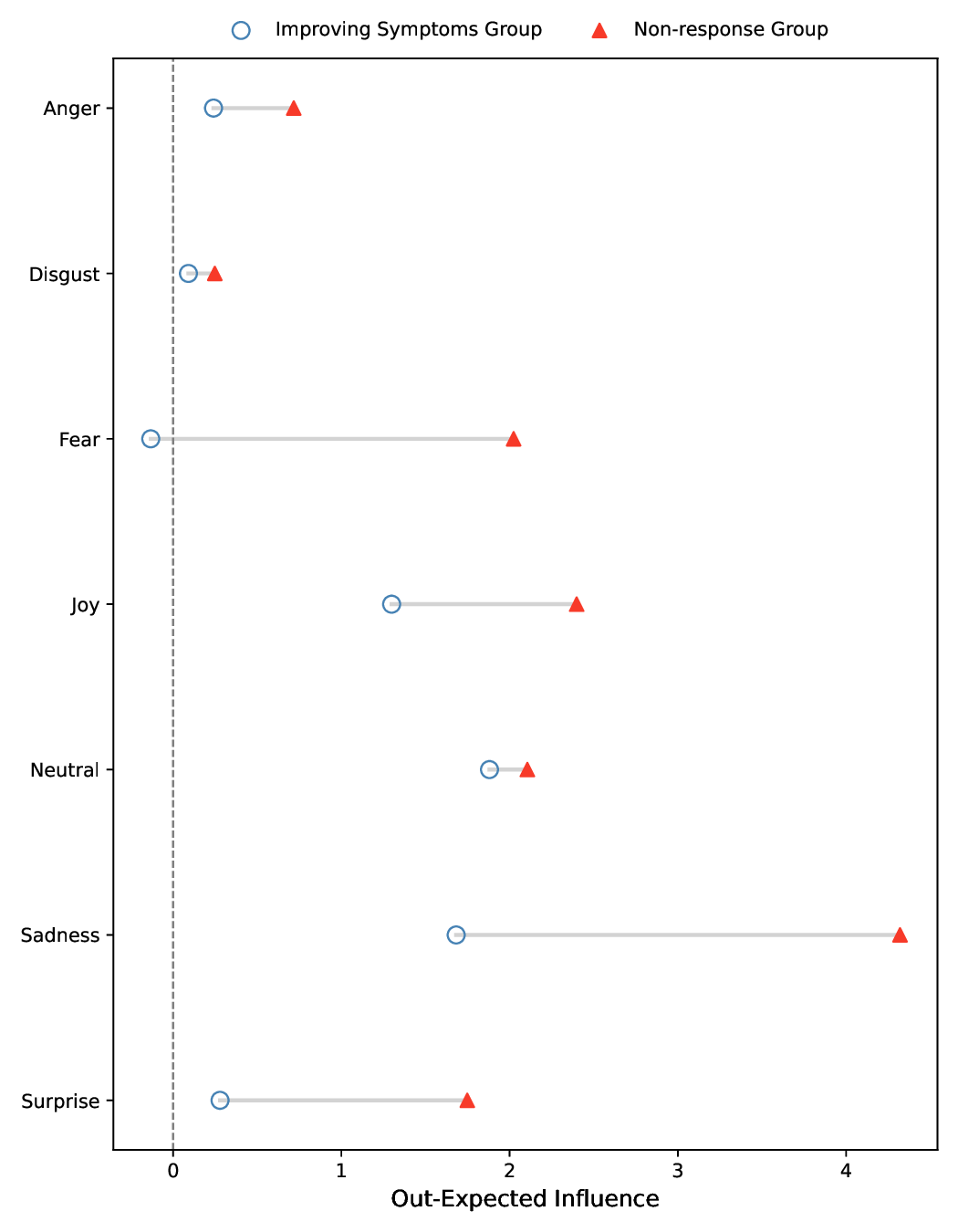}
    \caption{Out-Expected Influence in temporal network for both of the clusters returned by VISTA.}
    \label{fig:influence}
\end{figure}

To better characterize each emotion's effect on others, we examined out-expected influence (the degree to which each emotion influences the others) in Figure \ref{fig:influence}. The non-response group exhibits substantially higher out-expected influence for fear and sadness compared to the improving symptom group, which displays lower out-expected influence across all emotions. 

Combining the network analysis results with those from the regulatory flexibility framework, we observe how the improving symptoms group exhibits greater flexibility in expressing all emotions, evidenced by a lower out-expected influence for each emotion when compared to the non-response group. A higher out-expected influence for a given emotion, such as that of sadness and fear for the non-response group, indicates that an expression of that emotion could set off a cascade of expressions of other emotions. Within the non-response group, the higher out-expected influence of sadness and fear captures inflexibility as a marker of symptom non-improvement, as sadness has a notably higher influence on the expression of other emotions. In the improving symptom group, each symptom's lower influence on the others allows for an individual emotion to be expressed without a ``chain reaction.''

\section{Discussion}\label{sec:discussion}

Predicting treatment response in real-world mental health care remains a fundamental challenge, particularly when symptom assessments are sparse or inconsistently collected. Here, we demonstrate that passively captured emotional expressions in patient utterances, analyzed using a context-sensitive language model, can identify differential risk for treatment non-response. In a sample of over 12,000 psychotherapy patients we identified subgroups based on their expressed emotions, associated these clusters with clinical outcomes, and examined emotions dynamics using network analysis. 

Patients with improving symptoms exhibited flexible expression across the emotional spectrum, including joy, sadness, and neutral affect. In contrast, the non-response group displayed dynamics dominated by sadness and fear, with these negative emotions exerting disproportionate influence on subsequent affective states. These patterns align with the regulatory flexibility framework \cite{bonanno2013regulatory,bonanno2021resilience}: adaptive regulation involves balanced, context-sensitive expression, while inflexibility manifests as rigid, negatively-biased patterns. Notably, the proportion of patients classified as non-responders using these passive linguistic markers aligned with estimates derived from self-report measures in prior work \cite{pullmann2025message,hull2020two}, suggesting that non response can be characterized without requiring patient adherence to symptom surveys. 

Our approach combines extends prior work \cite{Shapira2022,Eberhardt2025,malgaroli2023natural,Burkhardt2021,Ewbank2021,Malgaroli2024,Nook2022} in three ways: capturing emotion at utterance-level granularity, linking temporal clusters to outcomes, and interpreting their emotion dynamics using network analysis. VISTA-SSM \cite{brindle2024vista} enables this by clustering irregularly sampled time series via stochastic modeling, parameterizing clusters with structural equations that can be used to generate networks and better understand their characteristics. Previous clinical studies \cite{coifman2007affect,heininga2019dynamical} suggested associations between emotion dynamics and clinical outcomes, but lacked the methodology to represent these dynamics as a network. Our method overcomes this hurdle and produces results consistent with research literature demonstrating associations between emotional flexibility and a clinically significant change in both anxiety and depression symptoms.\cite{kuppens2017emotion}. We make VISTA-SSM and our analytic scripts available
under an open-source license, to facilitate replication and extension to different data types.

Several limitations warrant consideration. First, our findings derive from message-based psychotherapy, which, although demonstrated to be equivalent in efficacy to face-to-face treatment \cite{pullmann2025message}, represents a specific therapeutic modality. Future work should examine replication of our findings to other delivery modalities. Future work could also address how our analysis was split into two distinct steps (clustering and network analysis) and attempt to combine these approaches. though this would detract from the applicability of VISTA-SSM by focusing the algorithm solely on interactions among covariates rather than the current structure which allows for the generation of time series from the distributions given by the estimated parameters. Furthermore, specific linguistic results in our sample may not be representative of larger populations or in languages other than English. It will be imperative, therefore, for future research using language models to include other languages to increase generalizability of findings. Another limitation is that our primary marker of emotional processes, emotional expression, is multiply determined \cite{lange2020toward}. Emotional expressions may represent the experience of bodily or subcortical states, for example, or appraisal of the ongoing context, schematic representation of broader situational factors, activated memories, attempts to influence the behavior of another person, or any combination of these factors \cite{ledoux2017higher}. Although emotional expressions in text are nonetheless heavily meaning laden and carry clear predictive value, future research should seek to include other emotional features. 

This study demonstrates that fine-grained emotional dynamics can serve as scalable, interpretable markers of treatment response, operating on passively collected data without patient burden. Critically, this approach requires no additional patient burden and can be implemented on existing clinical digital infrastructure. Emotion flexibility offers an interpretable framework for risk stratification that could identify patients needing intensified intervention before clinical deterioration. Future work should validate these findings across diverse therapeutic modalities, languages, and clinical populations to establish the generalizability of emotion-based digital phenotyping for precision mental health care.

\section{Declarations}

\subsection{Funding}
MM's research was supported by the National Institutes of Mental Health (NIMH) through grant K23MH134068. NC was supported by the National Science Foundation (NSF), award 2402555. The content is solely the responsibility of the authors and does not represent the official views of the NIMH or NSF.

\subsection{Competing Interests}
TDH is an employee of the platform that provided the data examined in this study. Talkspace had no role in the analysis, interpretation of the data, or decision to submit the manuscript for publication.

\subsection{Data Availability}
Requests for raw data can be made by written request to the corresponding author and require a separate Data Use Agreement.

\subsection{Code Availability}
Code may be found at \url{https://github.com/benjaminbrindle/vista_ssm}.

\subsection{Author's Contributions}\label{sec:contributions}
\textbf{BB}: Data Curation, Formal Analysis, Methodology, Software, Validation, Visualization, Writing – original draft, Writing – review and editing. \textbf{GAB}: Conceptualization, Validation,  Writing – original draft, Writing – review and editing. \textbf{TDH}: Data Curation, Project administration, Writing – original draft, Writing – review and editing. \textbf{NC}: Conceptualization, Methodology, Software, Supervision, Validation,  Writing – original draft, Writing – review and editing. \textbf{MM}: Conceptualization, Data Curation, Formal Analysis, Funding acquisition, Methodology, Project administration, Resources, Software, Supervision, Validation, Visualization, Writing – original draft, Writing – review and editing.

\section{Supplementary Material}

\setcounter{table}{0}
\renewcommand{\tablename}{Supplemental Table}

In supplemental table \ref{tab:symptom_pval}, we report the Bonferroni-corrected p-value \cite{bonferroni1936teoria} for each individual symptom, using the Mann-Whitney U test \cite{mann1947test} to determine if the distribution of symptoms in each of the clusters returned by VISTA differs, using both the first and last symptoms available for each patient during the study period. 

\begin{table}[h!]
    \centering
    \caption{Bonferroni-corrected p-values from Mann-Whitney U test on both initial (I) and final (F) symptoms for VISTA clusters (denoted C0 and C1), along with mean values for each cluster.}
    \begin{tabular}{p{0.11\textwidth}|p{0.10\textwidth}|p{0.10\textwidth}|p{0.10\textwidth}|p{0.10\textwidth}|p{0.10\textwidth}|p{0.10\textwidth}}
        Variable   & P-value (I) & P-value (F) & C0 Mean (I) & C1 Mean (I) & C0 Mean (F) & C1 Mean (F) \\ \hline
        Sleep      & 1.0000       & 1.0000         & 1.9394 & 1.9363 & 1.4609 & 1.4716 \\ \hline
        Weight     & 1.0000       & 1.0000         & 1.6470 & 1.6383 & 1.2012 & 1.1852 \\ \hline
        Concent    & 1.0000       & 1.0000         & 1.5571 & 1.5500 & 1.1002 & 1.1355 \\ \hline
        Mood       & 1.0000       & 1.0000         & 1.8174 & 1.8261 & 1.2267 & 1.2279 \\ \hline
        Fatigue    & 0.2044       & 1.0000         & 2.0158 & 2.0627 & 1.5742 & 1.5699 \\ \hline
        Anhed      & 1.0000       & 1.0000         & 1.6543 & 1.6449 & 1.1265 & 1.1186 \\ \hline
        Suicide    & 0.3702       & 1.0000         & 0.2767 & 0.2977 & 0.1884 & 0.2011 \\ \hline
        Worthl     & 0.0072       & 1.0000         & 1.9290 & 2.0008 & 1.2929 & 1.2874 \\ \hline
        Psychom    & 1.0000       & 1.0000         & 0.7173 & 0.6890 & 0.4609 & 0.4743 \\ \hline
        Nervous    & 0.6550       & 1.0000         & 2.3220 & 2.2914 & 1.6050 & 1.6295 \\ \hline
        UncWor     & 0.0005       & 1.0000         & 2.1537 & 2.0852 & 1.4022 & 1.4306 \\ \hline
        Fear       & 0.0007       & 1.0000         & 1.6538 & 1.5699 & 1.1158 & 1.1022 \\ \hline
        NoRelax    & 0.2213       & 1.0000         & 2.1074 & 2.0666 & 1.4504 & 1.4372 \\ \hline
        Irritab    & 0.0001       & 1.0000         & 1.7892 & 1.8736 & 1.3451 & 1.3246 \\ \hline
        GenWor     & 0.0028       & 1.0000         & 2.2406 & 2.1817 & 1.5233 & 1.5344 \\ \hline
        Restless   & 0.0366       & 1.0000         & 1.3469 & 1.2853 & 0.9143 & 0.9574 \\ \hline
    \end{tabular}
    \label{tab:symptom_pval}
\end{table}

\begin{table}[h!]
\centering
\caption{Slopes and intercepts by feature and cluster.}
\label{tab:vista}
\begin{tabular}{l|cc|cc}
& \multicolumn{2}{c}{VITA Cluster 0} & \multicolumn{2}{c}{VITA Cluster 1} \\
\hline
Emotion & Slope & Intercept & Slope & Intercept \\
\hline
Anger    & $-0.0021$ & $0.0599$  & $-0.00003$ & $0.1032$ \\
Disgust  & $-0.0011$ & $0.0308$  & $-0.0023$  & $0.0754$ \\
Fear     & $-0.0056$ & $0.1855$  & $-0.0031$  & $0.1758$ \\
Joy      & $0.0086$  & $0.1508$  & $0.0030$   & $0.1224$ \\
Neutral  & $-0.0076$ & $0.3168$  & $-0.0035$  & $0.2761$ \\
Sadness  & $0.0071$  & $0.1773$  & $0.0042$   & $0.1718$ \\
Surprise & $0.0007$  & $0.0791$  & $0.0017$   & $0.0759$ \\
\hline
\end{tabular}
\end{table}

\begin{table}[]
    \centering
    \caption{Results of logistic regression associating VISTA cluster membership with clinical outcomes while controlling for age, gender, and education.}

\begin{tabular}{l|l|l|l|l}
Clinical Outcome& Measure& Odds Ratio& 95\% CI& P\\ \hline
Significant Change& PHQ-9& 0.872 & 0.803, 0.947& .001\\ \hline
& GAD-7& 0.915 & 0.843, 0.994& .036\\ \hline
Response& PHQ-9& 0.812 & 0.746, 0.882& $<.001$\\ \hline
& GAD-7& 0.891 & 0.820, 0.969& .007\\ \hline
Remission& PHQ-9& 0.782 & 0.708, 0.864& $<.001$\\ \hline
& GAD-7& 0.919 & 0.822, 1.028& .139\\ \hline
Deterioration& PHQ-9& 1.127 & 1.026, 1.239& .013\\ \hline
& GAD-7& 1.145 & 1.036, 1.264& .008\\ \hline
\end{tabular}
\par\smallskip
\footnotesize\textit{Note.} Reference class is first cluster (Improving group).
\label{tab:logisticregression}
\end{table}

\newpage

Supplemental table \ref{tab:sup_demographics} shows covariates for the logistic regression of treatment outcomes by the clusters determined by the VISTA algorithm. Female gender was used as a reference class for the logistic regression analysis.

\begin{table}[h]
\centering
\caption{Demographics of patients (n=12,043) in our final sample, split by cluster: improving symptom group (n=8230) and non-response group (n=3813).}
\label{tab:sup_demographics}
\begin{tabular}{p{0.40\textwidth}p{0.25\textwidth}p{0.25\textwidth}}
\hline

Variable & \textbf{Improving Symptom Group} & \textbf{Non-response Group} \\
& n \% & n \% \\[1ex]
\hline

\textbf{Gender}                & \textbf{n = 6849}  & \textbf{n = 2999} \\
Female                         & 5300 (77.4)       & 2395 (79.9) \\
Male                           & 1436 (20.9)       & 550 (18.3) \\
Non-binary/other              & 113 (1.6)         & 54 (1.8) \\[1ex]

\textbf{Education Level}       & \textbf{n = 4030} & \textbf{n = 1868} \\
Bachelor Degree or Higher      & 2450 (60.8)       & 1133 (60.7) \\
High School                    & 611 (15.2)        & 289 (15.5) \\
Masters Degree                 & 333 (8.3)         & 146 (7.8) \\
Some College                   & 326 (8.1)         & 169 (9.0) \\
Associates Degree              & 139 (3.4)         & 55 (2.9) \\
Doctoral Degree                & 72 (1.8)          & 33 (1.8) \\
Professional Degree            & 54 (1.3)          & 23 (1.2) \\
Less than high school          & 45 (1.1)          & 20 (1.1) \\[1ex]

\textbf{Age Group}             & \textbf{n = 7758} & \textbf{n = 3547} \\
26-35                          & 4000 (51.6)       & 1757 (49.5) \\
18-25                          & 2225 (28.7)       & 1053 (29.7) \\
36-49                          & 1282 (16.5)       & 624 (17.6) \\
50+                            & 251 (3.2)         & 113 (3.2) \\

\hline
\end{tabular}
\end{table}

 \newpage

\bibliography{sn-bibliography}

\end{document}